\documentclass[runningheads]{llncs}

\usepackage{eccv}

\usepackage{eccvabbrv}
\usepackage{float}
\usepackage{arydshln}
\usepackage{tcolorbox}
\usepackage{enumitem}
\usepackage[dvipsnames]{xcolor}
\usepackage{graphicx}
\usepackage{lipsum}
\usepackage{booktabs}
\usepackage{multirow} 
\usepackage{tabulary}
\usepackage{wrapfig}
\usepackage{hyperref}
\usepackage{colortbl}
\usepackage[accsupp]{axessibility}  
\usepackage{listings}  
\usepackage{xcolor}     

\usepackage{hyperref}

\usepackage{orcidlink}

\begin{document}


\title{Path-level Hindsight Instructions for Semantic Exploration in Vision-Language Navigation} 

\titlerunning{$\Phi$-Nav}

\author{Sung June Kim\inst{1,2}$^\star$ \and
Sangpil Kim\inst{1}$^{\dagger}$ \and
Honglak Lee\inst{2}$^{\dagger}$
}
\authorrunning{S. J. Kim et al.}

\institute{Korea University, Seoul, South Korea \and
University of Michigan, Ann Arbor, Michigan, USA \\
}

\begingroup
\renewcommand\thefootnote{}
\footnotetext{\noindent $\star$ Work done at the University of Michigan as a visiting researcher.}
\addtocounter{footnote}{-1}
\endgroup

\begingroup
\renewcommand\thefootnote{}
\footnotetext{$\dagger$ Corresponing authors \{spk7@korea.ac.kr; honglak@umich.edu\}}
\addtocounter{footnote}{-1}
\endgroup

\maketitle
\setcounter{footnote}{0}
\begin{abstract}
On-policy exploration is a crucial component for training robust Vision-Language Navigation (VLN) agents, as it exposes the policy to a broader state distribution. However, such exploration inevitably leads to trajectories that deviate from expert demonstrations, resulting in a semantic mismatch between the executed visual stream and the original language instruction. In this work, we address this challenge by introducing $\Phi$-Nav, a unified on-policy framework that leverages hindsight reasoning to align instructions with the agent's actual exploratory journey.
Specifically, $\Phi$-Nav operates through a three-stage dual-supervision cycle: 1) the agent performs oracle-guided on-policy exploration, sampling a trajectory while learning from expert action feedback, 2) a hindsight speaker synthesizes a path-level hindsight instruction grounded in the collected visual observations, and 3) the agent conducts a second imitation pass, treating the synthesized trajectory–instruction pair as an additional expert demonstration. Through this process, $\Phi$-Nav bridges the critical semantic supervision gap inherent in on-policy methods, transforming semantically unlabeled movement into dense training signals. Evaluations on the R2R-CE and RxR-CE benchmarks show that $\Phi$-Nav yields competitive performance while requiring only a fraction of the expert demonstrations used by current baselines. These results underscore the necessity of semantic exploration in VLN, positioning $\Phi$-Nav as a effective solution for training embodied agents with limited data.

\end{abstract}

\section{Introduction}
\label{sec:intro}

The field of Vision-Language Navigation (VLN) is a cornerstone of embodied AI, requiring agents to ground natural language instructions in complex, previously unseen environments~\cite{anderson2018vision,wu2024vision}. To achieve robust generalization, modern VLN systems increasingly rely on on-policy exploration during training, exposing agents to diverse and dynamically evolving state distributions~\cite{liu2024vision,an2024etpnav,cheng2024navila,lee2024learning,zeng2025janusvln}. While architectural advances have significantly improved perception and cross-modal reasoning, the fundamental challenge of aligning exploratory behavior with coherent linguistic supervision remains unresolved.

To encourage robustness, many VLN policies adopt on-policy strategies such as scheduled sampling~\cite{bengio2015scheduled} and DAgger~\cite{ross2011reduction}, which mitigate exposure bias by allowing agents to act under their own policies while receiving expert action feedback~\footnote{In VLN, Scheduled Sampling and DAgger are often used interchangeably to describe online expert querying with a decaying student-action probability. Strictly, DAgger aggregates data with iterative retraining, whereas Scheduled Sampling performs on-the-fly policy mixing without dataset aggregation. In this work, we use the latter to denote the online variant of DAgger.}. Although these methods broaden the explored state space, they remain semantically tethered to static, pre-collected instructions. For example, when an agent deviates from the expert’s intended route, the original instruction no longer faithfully describes the executed visual stream. As a result, a substantial portion of exploratory trajectories lacks semantically aligned linguistic supervision, thereby limiting the full utilization of on-policy trajectories.

\begin{figure*}[t!]
\centering
\includegraphics[width=\linewidth]{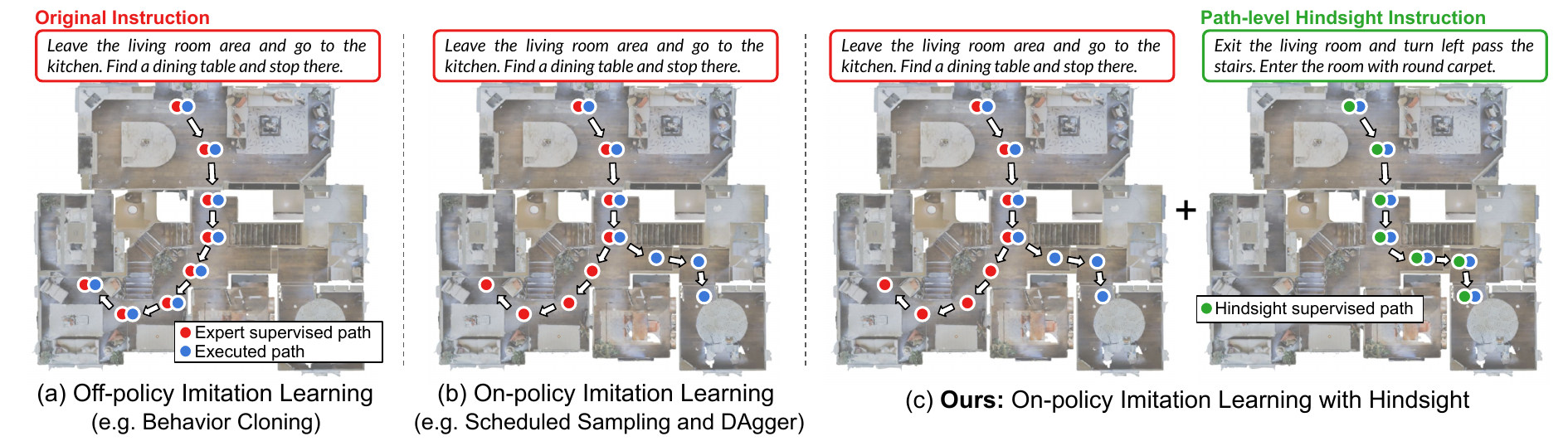}
\caption{\textbf{Conceptual comparison of VLN training strategies.} (a) Off-policy IL relies on static expert paths, failing to account for agent exploration. (b) On-policy IL exposes the agent to exploratory states but remains semantically limited to the original instruction, creating a supervision mismatch. (c) We utilize path-level hindsight relabeling to generate new instructions for exploratory trajectories, transforming deviations into meaningful training signals.}
\label{fig:pipeline}
\vspace{-2.0em}
\end{figure*}

To bridge the semantic supervision gap in on-policy VLN policy training, we propose $\Phi$(phi)-Nav, a framework that learns from path-level hindsight instructions. Our approach recognizes that every exploratory rollout contains a learnable narrative that remains invisible to traditional training objectives. To capture this latent knowledge, $\Phi$-Nav introduces a three-stage dual-supervision cycle that functions as a wrapper around existing on-policy algorithms. First, the agent samples a trajectory while receiving real-time action corrections from an expert oracle. Next, $\Phi$-Nav utilizes the robust multimodal spatio-temporal reasoning capability of pre-trained large vision-language models~(LVLM)~\cite{hurst2024gpt, bai2025qwen3} to retrospectively relabel its exploratory journey. By synthesizing path-level hindsight instructions that precisely describe the agent's actual visual observations, $\Phi$-Nav transform arbitrary exploratory noise into dense, linguistically-grounded training demonstrations. Lastly, the agent performs a second imitation pass, treating the synthesized trajectory–instruction pair as an additional expert demonstration, thereby reinforcing semantic grounding along the executed path.

Implementing hindsight relabeling for VLN presents unique challenges: unlike traditional hindsight paradigms that typically relabel final states within predefined goal space~\cite{andrychowicz2017hindsight}, our framework must narrate temporally-extended, open-ended trajectories, while at the same time mitigating LVLM hallucinations~\cite{li2023evaluating,wu2025generate}. We address these hurdles through two key mechanisms. First, we propose expert-in-context learning~\cite{brown2020language} to ensure synthesized instructions maintain structural and stylistic consistency with the training distribution. This enforces distributional alignment in linguistic style while allowing the specific visual observation to dictate the semantic content. Second, we introduce a trajectory–instruction alignment weighting mechanism~\cite{shi2022emscore, hessel2021clipscore, sarto2023positive} that adaptively weights hindsight supervision according to semantic fidelity. By evaluating the consistency between the visual trajectory and the synthesized instruction, the framework suppresses hallucinated or weakly grounded signals, ensuring that only reliable linguistic supervision influences policy optimization. Together, these components transform $\Phi$-Nav into self-correcting, autonomous supervision engine that significantly enhances sample efficiency.

We conduct extensive experiments on the R2R-CE and RxR-CE dataset, and demonstrate distinctive benefits of $\Phi$-Nav. Specifically, $\Phi$-Nav not only improves the performance of current on-policy training when utilizing the full expert dataset, but also remains highly competitive while requiring fewer expert demonstrations. This advantage suggests that our path-level hindsight instructions provide a richer supervision signal than traditional fixed-trajectory demonstrations alone, paving the way for autonomous agents that can effectively self-supervise and scale their intelligence through direct environmental experience.

The contributions of this work can be summarized as follows:
\begin{itemize}
    \item We propose a novel on-policy VLN framework $\Phi$-Nav that bridges the missing semantic supervision gap in training by converting exploratory rollouts into dense path-level hindsight instructions.
    \item We extend hindsight paradigms to temporally-continuous trajectories, utilizing expert-in-context learning for distributional alignment and trajectory-instruction alignment weighting to ensure accurate, hallucination-free hindsight supervision.
    \item Evaluations on R2R-CE and RxR-CE demonstrate that $\Phi$-Nav is sample-efficient, achieving competitive navigation performance while reducing dependence on expert data.
\end{itemize}

\section{Related Work}
\label{sec:related}

\subsection{Vision-Language Navigation}
Vision-Language Navigation~(VLN) is a fundamental task in embodied AI, requiring agents to ground and follow natural language instructions in 3D environments~\cite{anderson2018vision,wu2024vision, song2025towards, ko2025active, kim2025test, chen2025constraint}. The sequential decision making nature of the VLN task opts for policy training strategies such as reinforcement learning~(RL) or imitation learning~(IL). RL-based methods~\cite{wang2019reinforced,chen2021history, bundele2024scaling, zhang2025activevln, qi2025vln} optimize navigation policies by learning to maximize a predefined reward/score function. 
However, due to the innate difficulties of RL in reward modeling, many VLN policies utilize IL as their main objectives~\cite{anderson2018vision, krantz2020beyond,chen2022think, an2023bevbert, an2024etpnav, kamath2023new,gao20253d, wei2025streamvln}. These methods utilize on-policy trajectory sampling methods such as scheduled sampling~\cite{bengio2015scheduled} and DAgger~\cite{ross2011reduction} since they effectively mitigate exposure bias by forcing the agent to learn from its own mistakes, thereby narrowing the distribution shift between training and inference. For instance, the foundational approach in \cite{anderson2018vision} introduced student-forcing to the VLN domain, demonstrating that training on the agent's own sampled actions is essential for learning error-recovery behaviors. 

Despite these advancements, a fundamental bottleneck remains: when an agent explores paths far from the expert demonstration, the original instruction becomes semantically irrelevant, leading to a semantic supervision gap. Our proposed $\Phi$-Nav bridges this gap by generating path-level hindsight instructions that provide precise, dense supervision for any exploratory action the agent executes, effectively scaling training beyond the limits of static expert data.
\vspace{-1em}

\subsection{Navigation Instruction Generation}
Navigation Instruction Generation is an important sub-task in the VLN research, which enables precise and interpretable human-robot interaction. The Speaker-Follower framework~\cite{fried2018speaker} pioneered the "Speaker" model to back-translate paths into instructions, allowing for the creation of vast amounts of synthetic training data. This concept was further refined by EnvDrop~\cite{tan2019learning}, which performs environmental dropout to synthesize navigation pairs and significantly improve generalization. Marky~\cite{wang2022less} further scaled the augmentation scheme by extracting visual landmarks. Recent works leverage the vast pretrained knowledge of LVLMs, either utilizing them as a zero-shot generator~\cite{yan2024instrugen,zheng2026canespeaker} or further fine-tuning them to generate linguistically diverse and spatially grounded instructions~\cite{fan2025scene,kong2024controllable,fan2024navigation}. 

Generating hindsight instructions as in $\Phi$-Nav poses several distinct challenges that separate its purpose from these works. While previous works focus on pretraining or offline augmentation, our framework is designed for providing semantic supervision for exploratory paths within on-policy training stream, which is a overlooked paradigm in existing literature. Furthermore, since on-policy exploration often yields suboptimal loops or deviations, $\Phi$-Nav must rigorously verify the instruction stream. This ensures that only semantically- and logically-consistent labels are utilized to supervise the navigation policy.
\vspace{-1em}

\subsection{Hindsight Experience Learning}

Hindsight experience learning was introduced to improve sample efficiency in sparse-reward settings by relabeling failed trajectories with alternative achieved goals. The seminal Hindsight Experience Replay (HER)~\cite{andrychowicz2017hindsight} introduced goal substitution to improve sample efficiency. Subsequent works extended hindsight relabeling to visual goal-conditioned settings~\cite{nair2018visual,pong2018temporal}, hierarchical/multi-goal RL~\cite{levy2019hierarchical,nachum2018data}, model-based imagination~\cite{kaiser2020model,hafner2020dream}, and curriculum/prioritized relabeling~\cite{fang2019curriculum,luo2020energy}. More recently, inspired by hindsight principles, language models have been used to provide retrospective feedback and self-refinement signals for policy learning~\cite{huang2022language,zhou2023language,xu2023reflexion,liang2023react}.

THER~\cite{cideron2019self} and HSL~\cite{li2026spinning} are closely related to our work in their use of language-guided hindsight signals, yet these methods mainly operate over structured predefined goal spaces, typically relabeling terminal states or scalar rewards. In contrast, $\Phi$-Nav extends hindsight learning from goal substitution to temporally coherent, fine-grained path-level instruction relabeling. By integrating LVLM-based instruction synthesis with semantic verification, our framework establishes a hindsight paradigm tailored to embodied language grounding in on-policy VLN training.


\section{Method}
\label{sec:method}
\begin{figure*}[t!]
\centering
\includegraphics[width=\linewidth]{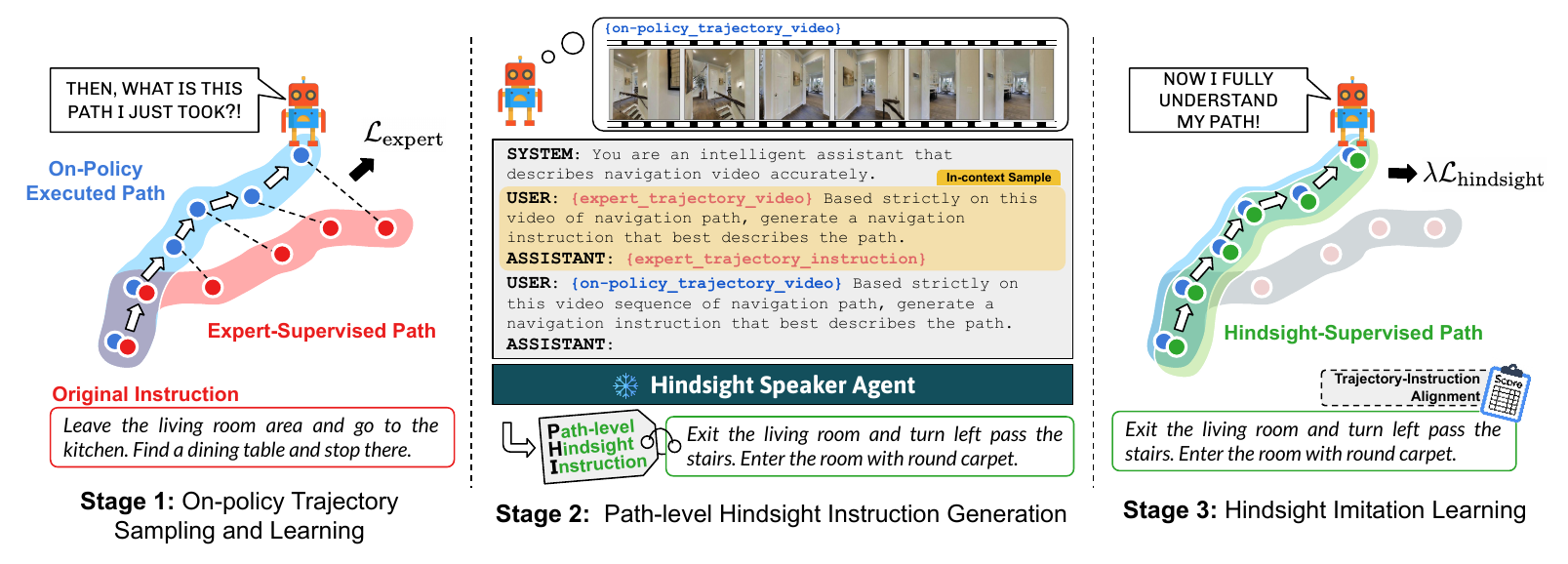}
\caption{\textbf{Overview of the $\Phi$-Nav pipeline.} In Stage 1, the agent samples a trajectory while being supervised by expert actions. In Stage 2, the agent analyzes its path in hindsight to generate a verified instruction that aligns with the observed visual stream. Lastly, in Stage 3, the agent re-evaluates its past decision points under the guidance of this new instruction, learning to ground its actions with semantic supervision that aligns. Full prompt is presented in the Appendix.}
\label{fig:overview}
\end{figure*}

In this section, we detail the architecture and training paradigm of $\Phi$-Nav. We first define the navigation environment and our exploration strategy in Section \ref{sec:prob}. Next, we detail the three stages of our pipeline: on-policy trajectory sampling (Section \ref{sec:s1}), path-level hindsight instruction generation (Section \ref{sec:s2}), and the final dual-supervision optimization process (Section \ref{sec:s3}). The overall framework of $\Phi$-Nav is illustrated in Figure \ref{fig:overview}.

\subsection{Problem Formulation}
\label{sec:prob}

\subsubsection{Vision-Language Navigation.}
In the Vision-Language Navigation (VLN) task, the agent must traverse a 3D environment to reach a destination specified by a natural language instruction $I$. At each discrete time step $t$, the agent receives a visual observation $v_t$, either in monocular or panoramic view. Based on the current visual observation and the instruction, the agent predicts next action $a_t\in \mathcal{A}$, where $\mathcal{A}$ is the total action space. The primary objective is to learn a policy $\pi_\theta(a_t | v_t, I)$ that maximizes the ratio of successfully reaching the goal while ensuring that every navigational action is precisely grounded in the semantic constraints of the instruction.

\subsubsection{On-policy Exploration Strategies.}
To enable robust navigation, VLN policies frequently employ on-policy exploration strategies such as Scheduled Sampling~\cite{bengio2015scheduled} and DAgger~\cite{ross2011reduction} to expose the agent to a broader range of the state space. At each time step $t$, the action $a_t$ is sampled according to:
$$a_t \sim
\begin{cases}
a_t^* & \text{w. p. } \epsilon^{(i)} \\
 \pi_\theta(a | v_t, I^*_{E}) & \text{w. p. } 1 - \epsilon^{(i)},
\end{cases}$$
where $a_t^*$ denotes the expert action and $\epsilon^{(i)}$ is a decay factor for the $i$-th training episode that gradually shifts the sampling from the expert's teacher-forcing to the agent's current policy $\pi_\theta$. While these methods effectively expand the visited state space, they introduce a significant semantic gap. When the agent samples $a_t \neq a_t^*$, it generates an exploratory trajectory $\tau = \{v_0, a_0, \dots, v_T, a_T\}$ that deviates from the expert path $\tau^*_{E}$, making the instruction for the expert trajectory $I^*_{E}$ semantically unmatched with the resulting visual stream. To address these deviations, $\Phi$-Nav introduces a hindsight mechanism in VLN by synthesizing instructions that accurately reflect the agent’s actual exploratory paths. By providing valid linguistic supervision for the path actually executed, $\Phi$-Nav bridges the gap between exploration and semantic instruction grounding.

\subsection{Stage 1: On-policy Trajectory Sampling and Learning}
\label{sec:s1}
$\Phi$-Nav begins with the standard training procedure adopted by prior VLN policies~\cite{chen2022think, an2024etpnav, krantz2020beyond}. At each time step $t$, the agent processes its visual observation and the original language instruction to generate a probability distribution of possible actions using policy $\pi_\theta$. Then, following the sampling strategies in Section~\ref{sec:prob}, the navigation agent initiates exploration to collect on-policy trajectory $\tau = \{v_0, a_0, \dots, v_T, a_T\}$. Regardless of the selected actions, the agent's own predictions are supervised and corrected by the offline expert action $a^{\ast}$, by learning to minimize the cross-entropy loss:
\begin{equation}
\label{eq:1}
    \mathcal{L}_{\text{expert}} = -\sum_{t=0}^T \log \pi_\theta(a_t^* \mid v_t, I^*_{E}).
\end{equation}
Through this supervision, the agent learns how to mimic the expert's decision-making process within the observed states. However, most works stop here, leaving the semantic supervision for exploratory, on-policy trajectories unaddressed, as the instruction for the expert trajectory $I^*_{E}$ may become misaligned with the agent’s executed trajectory. $\Phi$-Nav effectively bridges this gap through the following stages.

\subsection{Stage 2: Path-level Hindsight Instruction Generation}
\label{sec:s2}
The core of $\Phi$-Nav lies in its ability to transform raw, exploratory visual streams into semantically-grounded language instruction. Once a trajectory $\tau$ is buffered from Stage 1, we employ a large vision-language model~(LVLM) based Hindsight Speaker Agent with expert-in-context learning strategy to synthesize a reliable hindsight instruction $I_{H}$.

\subsubsection{Hindsight Speaker Agent. }
We leverage the extensive zero-shot spatio-temporal reasoning and linguistic knowledge of LVLMs to act as the Hindsight Speaker Agent~(HSA). The advantage of zero-shot LVLMs over task-specific fine-tuned speakers~\cite{fan2025scene,kong2024controllable} lies in their stronger generalization and linguistic flexibility under distribution shift. While fine-tuned speakers are typically trained to reconstruct instructions conditioned on ideal ground-truth paths, zero-shot HSA can generate visually grounded descriptions even when the agent substantially deviates from the expert route, which is a common scenario for on-policy trajectories. For instance, if the agent takes a wrong turn and traverses a visual state rarely observed  in training demonstrations, a fine-tuned speaker may struggle to produce a coherent or accurate instruction due to its reliance on expert-aligned data. In contrast, HSA can still generate a faithful description of the detour, thereby maintaining semantic alignment with the executed trajectory.

\subsubsection{Expert-In-Context Learning. }
To further stabilize hindsight instruction generation, we incorporate an expert-in-context learning strategy. Although HSA could operate in a zero-shot manner, we provide a single exemplar consisting of an expert trajectory and its corresponding human-annotated instruction. Specifically, for each generation, we sample a trajectory–instruction pair:
\[
(\tau^{\ast}, I^{\ast}) \sim p_{\mathcal{E}},
\]
where $p_{\mathcal{E}}$ denotes the empirical distribution induced by the offline expert demonstration set $\mathcal{E} = \{(\tau_i^{\ast}, I_i^{\ast})\}_{i=1}^{N}$.
Pure zero-shot generation may introduce excessive diversity or creativity in phrasing, which can induce distribution shift and hinder policy learning. By contrast, conditioning on an expert demonstration encourages HSA to produce instructions that resemble human-authored navigation commands in tone, structure, and granularity, while still grounding the semantic content in the visual observations of the current on-policy trajectory.

\subsection{Stage 3: Hindsight Imitation Learning}
\label{sec:s3}

Once the path-level hindsight instruction is generated, $\Phi$-Nav conducts a second round of imitation learning using the synthesized trajectory–instruction pair as augmented supervision. However, because the instruction is produced by an LVLM, it may contain imperfect or weakly grounded descriptions. To mitigate potential noise, we introduce a trajectory–instruction alignment score that weights the contribution of hindsight supervision during policy optimization.

\subsubsection{Trajectory-Instruction Alignment Weighting. }
\label{sec:s3-1}
Since $\Phi$-Nav operates within the on-policy training stream, the alignment between a trajectory and its hindsight instruction must be evaluated online and without ground-truth references. To this end, we design a lightweight trajectory–instruction alignment scoring module inspired by EMScore~\cite{shi2022emscore}, which evaluates video–caption consistency via hierarchical embedding similarities: a coarse score computed from global video and sentence embeddings, and a fine-grained score from frame–word similarity.

While this hierarchical design is well suited for generic video captioning, navigation scenarios place particular emphasis on landmark grounding, since they serve as critical spatial anchors for decision making. Therefore, instead of computing fine-grained similarity over all words, we focus specifically on landmark nouns in the instruction and measure their alignment with individual trajectory frames. First, we compute the coarse alignment score as a cosine similarity between global trajectory video and sentence embeddings:
\begin{equation}
    \mathcal{S}_\text{coarse} = \text{sim}\Big(\frac{1}{|V|}\sum_{v_{i}\in V}f(v_i),\ g(I_H)\Big),
\end{equation}
where $V$ denotes the whole video frames and $f$ and $g$ are the CLIP~\cite{radford2021learning} image and text encoder, respectively. Next, for fine-grained alignment, we first extract landmark nouns from $I_H$ using spaCy~\cite{spacy2} and obtain a set of landmark nouns ${M}$. We then compute bidirectional frame–landmark alignment scores:
\begin{equation}
    A_{M \xrightarrow{}V}=\frac{1}{|M|}\sum_{m_{j}\in M}\max_{v_i\in V}\text{sim}\big(f(v_i),\ g(m_{j})\big)
\end{equation}
\begin{equation}
    A_{V \xrightarrow{}M}=\frac{1}{|V|}\sum_{v_{i}\in V}\max_{m_j\in M}\text{sim}\big(f(v_i),\ g(m_{j})\big),
\end{equation}
and define the fine-grained score as their harmonic mean:
\begin{equation}
\mathcal{S}_\text{fine}=\frac{2 \cdot A^{+}_{M \xrightarrow{}V} \cdot A^{+}_{V \xrightarrow{}M}}{A^{+}_{M \xrightarrow{}V} + A^{+}_{V \xrightarrow{}M}+\epsilon},
\end{equation}
where $A^+=\max(0,A)$ ensures non-negative alignment contributions, and $\epsilon>0$ is a small constant to prevent division by zero. We define the trajectory-instruction alignment weight $\lambda$ as the average of $\mathcal{S}_\text{coarse}$ and $\mathcal{S}_\text{fine}$, which adaptively modulates the contribution of hindsight signals in each episode. While more sophisticated online reference-free scoring strategies may be explored, we adopt this lightweight formulation for efficiency and leave further improvements to future work.

\subsubsection{Hindsight Loss Function. }
In the final stage of the $\Phi$-Nav cycle, the agent updates its policy by treating its own exploratory journey as a successful execution of the newly synthesized instruction. Given a completed on-policy trajectory $\tau = \{v_0, a_0, \dots, v_T, a_T\}$ and its corresponding hindsight instruction $I_H$, the agent is supervised to maximize the likelihood of the actions it actually performed. The hindsight imitation loss is formulated as:
\begin{equation}
    \mathcal{L}_{\text{hindsight}} = -\sum_{t=0}^T \log \pi_\theta(a_t \mid v_t, I_{H}).
\end{equation}
Crucially, this objective allows the agent to learn the relationship between linguistic cues and visual state transitions even in regions of the environment far from the expert distribution. To balance traditional imitation learning with this self-supervised signal, we combine the expert loss from Eq.~\ref{eq:1} with the weighted hindsight loss. The total optimization objective is defined as:
\begin{equation}
    \mathcal{L}=\mathcal{L}_{\text{expert}} + \lambda \mathcal{L}_{\text{hindsight}}
\end{equation}
where $\lambda$ is the weighting coefficient derived from the trajectory–instruction alignment scores. Through this optimization, $\Phi$-Nav ensures that every navigation episode, whether successful or exploratory, contributes to a more robust and semantically-aware on-policy training.

\section{Experiment}
\label{sec:exp}

\subsection{Experiment Setup}
\subsubsection{Datasets. }
$\Phi$-Nav is trained and evaluated on the R2R-CE and RxR-CE datasets. These datasets provide expert navigation trajectories paired with natural language instructions. The visual observations along each trajectory are rendered using the Habitat Simulator~\cite{savva2019habitat}, enabling photo-realistic embodied navigation in continuous 3D environments. The R2R-CE dataset consists of 5,611 trajectories divided into train, validation seen, validation unseen, and test unseen~\footnote{Due to the deprecation of the official VLN-CE evaluation server on \href{https://eval.ai/web/challenges/challenge-page/719/overview}{Eval.ai}, we were unable to obtain test-unseen results.}. Each trajectory is annotated with three English instructions. The dataset features an average path length of 9.89 meters and an average instruction length of 32 words. The RxR-CE dataset provides navigation instructions in three languages—English, Hindi, and Telugu—with substantially longer descriptions averaging approximately 120 words per instruction. In addition, RxR-CE features longer trajectories on average than R2R-CE, increasing the difficulty of long-horizon reasoning and fine-grained instruction grounding.

\subsubsection{Evaluation Metrics. }
For navigation performances, we adopt the following metrics. Navigation Error (NE): average distance in metric space between the final and target
location; Success Rate (SR): the ratio of episodes with NE less than 3.0 meters; Oracle SR (OSR): SR given the oracle stop policy; SR penalized by Path Length (SPL), Trajectory Length (TL): average path length in meters; Normalize Dynamic Time Wrapping (NDTW): the similarity between the predicted and expert paths and lastly NDTW penalized by SR (SDTW). For instruction generation, we report the BLEU~\cite{papineni2002bleu} and ROUGE-L~\cite{lin2004rouge} metric in Section~\ref{sec:insgen},which quantify lexical overlap and sequence-level similarity between generated instructions and the ground-truth references.

\subsubsection{Baselines and Implementation Details. }
We evaluate the applicability of $\Phi$-Nav under two on-policy imitation learning paradigms: DAgger~\cite{ross2011reduction} and scheduled sampling~\cite{bengio2015scheduled}. For the DAgger setting, we adopt CMA~\cite{krantz2020beyond} as the baseline policy. CMA is a representative baseline in the VLN-CE benchmark, designed to model multimodal grounding via cross-modal attention and temporal dependencies through recurrent neural networks. Specifically, we use the CMA-DA and CMA-DA-PM-Aug variants, which are the DAgger-trained versions of CMA. For these experiments, we follow the DAgger sample collection strategy and store the hindsight-generated trajectory–instruction pairs in the same buffer for subsequent training. Next, for the scheduled sampling setting, we adopt ETPNav~\cite{an2024etpnav}, which leverages a topological map for structured decision-making. Here, the policy updates are performed on a per-sample basis. Accordingly, $\Phi$-Nav is applied in an iterative three-stage cycle for each sampled trajectory.

The Hindsight Speaker Agent is implemented using the Qwen2.5-VL-7B model~\cite{bai2025qwen3} as backbone. We provide ablation on different backbone choices in Section~\ref{sec:ablation}. Additionally, we provide full prompt template and hyperparameters in the Appendix. For panoramic observation settings, we select four non-overlapping views centered around the front-facing direction and use them as input frames to the Hindsight Speaker Agent. The rest of the training configuration adhere to the default setting of the baseline policies. The experiments were conducted on NVIDIA L40S GPUs.

\subsection{Main Navigation Results}

\subsubsection{Evaluation on R2R-CE. }
The results in Table~\ref{tab:r2r-ce} indicate that incorporating additional semantic exploration through $\Phi$-Nav consistently improves performance under on-policy training paradigms. For example, in DAgger-based experiments, $\Phi$-Nav improves the SR of the CMA-D-P-A baseline by 1.98 percentage points on the val unseen split. Similarly, in scheduled sampling based experiments, $\Phi$-Nav improves the SR and the SPL of ETPNav by 5.37 and 3.59 percentage points in the val unseen split, respectively. It is also worth noting that although ETPNav alone achieves lower performance, integrating $\Phi$-Nav improves it to slightly surpass the recent state-of-the-art method g3D-LF~\cite{wang2025g3d}. These results clearly highlight the necessity of filling the missing semantic supervision gap in on-policy explorations.

\subsubsection{Evaluation on RxR-CE. }
Consistent with the results on R2R-CE, Table~\ref{tab:rxr-ce} shows that $\Phi$-Nav achieves stable numerical improvements across both validation splits. In particular, on the val unseen split, $\Phi$-Nav improves SR and SPL by 1.04 and 1.06 percentage points, respectively. Additionally, gains in the DTW-based metrics indicate improved trajectory fidelity even in the linguistically dense and long-horizon navigation scenarios of RxR-CE. However, the magnitude of improvement is smaller than on R2R-CE, likely because RxR-CE requires generating finer-grained instructions over substantially longer trajectories, making accurate hindsight instruction generation more challenging.

\begin{table*}[t]
	\centering
	\caption{Experimental results on the R2R-CE dataset. Performance improvements obtained by applying $\Phi$-Nav are highlighted in bold. Methods marked with $\dagger$ use monocular-view observations.}
    \vspace{-0.75em}
\label{tab:r2r-ce}
\resizebox{0.98\textwidth}{!}{
	\begin{tabular}{*{2}{r}|*{5}{c}|*{5}{c}}
		\toprule
            &
		\multicolumn{1}{l|}{\textbf{Methods}}
            & \multicolumn{5}{c}{Val Seen} 
            & \multicolumn{5}{c}{Val Unseen} \\

		\midrule

        &\multicolumn{1}{l|}{\cellcolor{blue!8} \textbf{DAgger}~\cite{ross2011reduction}}
        & TL & NE$\downarrow$ & OSR$\uparrow$ & SR$\uparrow$ &  SPL$\uparrow$ 
		& TL & NE$\downarrow$ & OSR$\uparrow$ & SR$\uparrow$ &  SPL$\uparrow$  \\
        \midrule
        
        &  CMA-D$^{\dagger}$~\cite{krantz2020beyond}
		& 8.64 & 6.79 & 41.77 & 32.87 & 30.28
        & 7.86 & 8.17 & 33.22 & 26.64 & 24.88  \\
            
        &  \multicolumn{1}{r|}{\cellcolor{gray!20} w/ $\Phi$-Nav\textbf{}}
        & \cellcolor{gray!20}8.32 & \cellcolor{gray!20}\textbf{6.71} &  \cellcolor{gray!20}\textbf{43.44} & \cellcolor{gray!20}\textbf{34.83} & \cellcolor{gray!20}\textbf{33.13} 
        & \cellcolor{gray!20}7.89 & \cellcolor{gray!20}\textbf{7.69} & \cellcolor{gray!20}\textbf{35.12} & \cellcolor{gray!20}\textbf{27.62} & \cellcolor{gray!20}\textbf{25.92}\\

        &  CMA-D-P-A$^{\dagger}$~\cite{krantz2020beyond}
		  & 9.63 & 7.11&46.13 & 37.17 & 34.82  
        & 8.86 & 7.62 & 40.28 & 32.10 & 29.92\\
            
        &  \multicolumn{1}{r|}{\cellcolor{gray!20} w/ $\Phi$-Nav\textbf{}}
        & \cellcolor{gray!20}9.54 & \cellcolor{gray!20}\textbf{6.99} &  \cellcolor{gray!20}\textbf{47.18} & \cellcolor{gray!20}\textbf{38.80} & \cellcolor{gray!20}\textbf{35.63} 
        & \cellcolor{gray!20}8.27 & \cellcolor{gray!20}\textbf{7.42} & \cellcolor{gray!20}\textbf{42.23} & \cellcolor{gray!20}\textbf{34.08} & \cellcolor{gray!20}\textbf{31.51}\\

          \midrule
          \midrule

        &\multicolumn{1}{l|}{\cellcolor{blue!8} \textbf{Sched. Sampling}~\cite{bengio2015scheduled}}
        & TL & NE$\downarrow$ & OSR$\uparrow$ & SR$\uparrow$ &  SPL$\uparrow$ 
		& TL & NE$\downarrow$ & OSR$\uparrow$ & SR$\uparrow$ &  SPL$\uparrow$  \\
        \midrule
 
        & VLN-BERT~\cite{hong2022bridging}
		  &12.50 & 5.02&  59 &50 &44
        & 12.23& 5.74& 53 & 44 &39  \\
        
        & BEVBert~\cite{an2023bevbert}
		& 13.98 & 3.77 & 73 & 68 & 60  
        & - & 4.57& 67 & 59 & 50\\

        & HNR~\cite{wang2024lookahead}
		&11.79&3.67& 76& 69& 61
        &12.64&4.42& 67 &61& 51 \\

		& ENP-ETPNav~\cite{liu2024vision}
		  & 11.82   & 3.90& 73 & 68 & 59
        & 11.45& 4.69 & 65 & 58 & 50\\
  
		& g3D-LF~\cite{wang2025g3d}
		  & 11.61 & 3.72 & 72 & 63 & 55
        & -  & 4.53 & 68 & 61 & 52\\
        \midrule

        & ETPNav~\cite{an2024etpnav}
		  & 11.38 & 3.94& 72.23 & 66.45 & 59.62
        & 11.99& 4.71 & 64.71 & 57.21 & 49.15\\
             
        &  \multicolumn{1}{r|}{\cellcolor{gray!20} w/ $\Phi$-Nav\textbf{}}
         & \cellcolor{gray!20}11.29 & \cellcolor{gray!20}\textbf{3.38} &  \cellcolor{gray!20}\textbf{75.06} & \cellcolor{gray!20}\textbf{68.63} & \cellcolor{gray!20}\textbf{61.41} 
         & \cellcolor{gray!20}12.76 & \cellcolor{gray!20}\textbf{4.29} & \cellcolor{gray!20}\textbf{68.84} & \cellcolor{gray!20}\textbf{62.58} & \cellcolor{gray!20}\textbf{52.74}\\
		  
		\bottomrule
	\end{tabular}
}
\end{table*}

\begin{table*}[t]
	\centering
	\caption{Experimental results on the RxR-CE dataset. Performance improvements obtained by applying $\Phi$-Nav are highlighted in bold.}
    \vspace{-0.75em}
    \label{tab:rxr-ce}
    \resizebox{0.98\textwidth}{!}{
	\begin{tabular}{*{2}{r}|*{5}{c}|*{5}{c}}
		\toprule
            &
		\multicolumn{1}{l|}{\textbf{Methods}}
            & \multicolumn{5}{c}{Val Seen} 
            & \multicolumn{5}{c}{Val Unseen} \\

		\midrule

        &\multicolumn{1}{l|}{\cellcolor{blue!8} \textbf{Sched. Sampling}~\cite{bengio2015scheduled}}
        & NE$\downarrow$ & SR$\uparrow$ &  SPL$\uparrow$ & NDTW$\uparrow$ & SDTW$\uparrow$
		& NE$\downarrow$ & SR$\uparrow$ &  SPL$\uparrow$ & NDTW$\uparrow$ & SDTW$\uparrow$ \\
        \midrule

        & VLN-BERT~\cite{hong2022bridging}
		  &- &- &- &- &- 
        &8.98 &27.08 &22.65 &46.71 &- \\

        & HNR~\cite{wang2024lookahead}
		  &4.85 &63.72 &53.17 &68.81 &52.78 
        &5.51 &56.39 &46.73 &63.56 &47.24 \\

		& ENP-ETPNav~\cite{liu2024vision}
		&5.10 &62.01 &51.18 &67.22 &51.90 
        &5.51 &55.27 &45.11 &62.97 &45.83 \\

        \midrule
        & ETPNav~\cite{an2024etpnav}
		& 5.03 & 61.46 & 50.83 & 66.41 & 51.28 
        & 5.64 & 54.79 & 44.89 & 61.90 & 45.33 \\
             
        &  \multicolumn{1}{r|}{\cellcolor{gray!20} w/ $\Phi$-Nav\textbf{}}
		&  \cellcolor{gray!20}\textbf{5.02} & \cellcolor{gray!20}\textbf{62.84} & \cellcolor{gray!20}\textbf{51.34} & \cellcolor{gray!20}\textbf{67.92} & \cellcolor{gray!20}\textbf{51.70}
        & \cellcolor{gray!20}5.67 & \cellcolor{gray!20}\textbf{55.83} & \cellcolor{gray!20}\textbf{45.95} & \cellcolor{gray!20}\textbf{62.93} & \cellcolor{gray!20}\textbf{46.11} \\

		\bottomrule
	\end{tabular}
}
\vspace{-1.0em}
\end{table*}


\subsection{Instruction Generation Analysis}
\label{sec:insgen}
\subsubsection{Trajectory-Instruction Alignment. }
In this section, we visually examine how the generated instructions are semantically grounded in the visual stream of the trajectories and how such grounding contributes to hindsight imitation learning, as quantified by the trajectory–instruction alignment weight (TIAW)~\footnote{The distribution of TIAW scores across trajectories is provided in the Appendix.}. In the first example of Figure~\ref{fig:instructions}, the on-policy trajectory diverges from the expert path at an early stage. However, $\Phi$-Nav accurately captures the actual executed trajectory, introducing additional descriptive phrase~(\textit{e.g., large mirror on the wall}) that encourages richer visual grounding along the way. The final position aligns precisely with the \textit{kitchen counter} stop signal, yielding a relatively high TIAW of $0.53$. In the second example, the generated instruction accurately captures directional cues (\textit{e.g., right, left}), as well as fine-grained visual details. Compared to the first example, however, many of the referenced landmarks occupy relatively small regions in the pixel space. This reduced perceptual saliency weakens the visual grounding signal, leading to slightly lower TIAW of $0.46$ despite the instruction remaining semantically accurate. In the final example, the observations lack distinctive landmarks, hence the visual stream provides limited object-level anchors for semantic grounding. Consequently, the trajectory–instruction correspondence is substantially weaker, yielding a low TIAW of 0.22. In summary, $\Phi$-Nav selectively emphasizes well-grounded trajectories and down-weights weakly aligned ones, enabling more stable and adaptive optimization.

\begin{figure*}[t!]
\centering
\includegraphics[width=\linewidth]{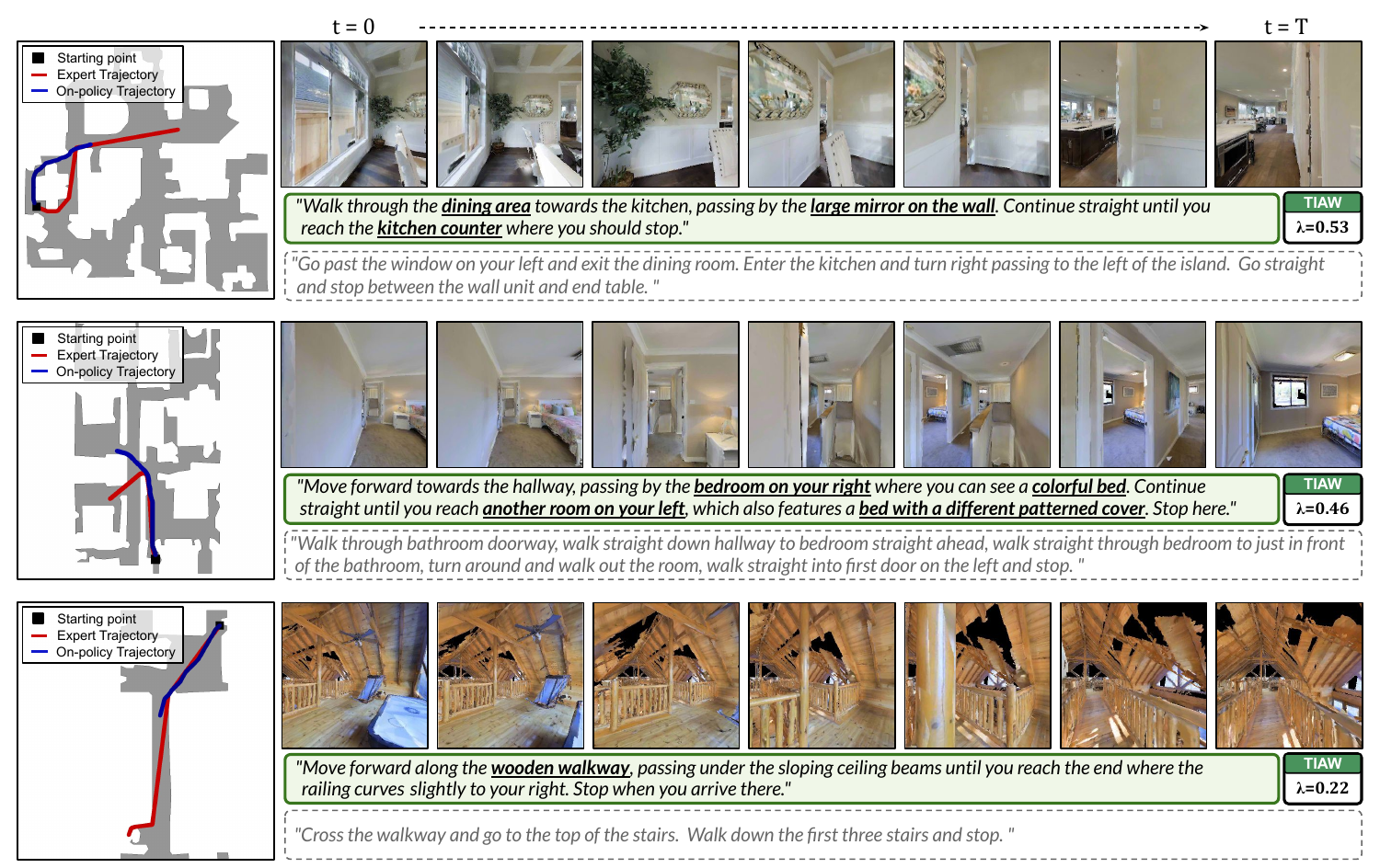}
\vspace{-1.5em}
\caption{\textbf{Trajectory–Instruction Alignment.} Generated hindsight instructions are shown in green boxes, with major landmarks highlighted in bold and underlined. The gray text below each corresponds to the original expert instruction.}
\vspace{-1.5em}
\label{fig:instructions}
\end{figure*}

\subsubsection{Effect of Expert-In-Context Learning. }
Our proposed expert-in-context learning strategy constrains the open-ended language space to remain within the training instruction distribution for stable hindsight optimization. We validate this by comparing against a pure zero-shot setting and a template-based prompting strategy adopted in InstruGen~\cite{yan2024instrugen}, generating instructions on offline expert trajectories and measuring their lexical and structural similarity to the ground-truth instructions of R2R-CE. For these comparisons, we exclude the original instruction from each trajectory to avoid trivial matches. We use POS-tagging~\cite{spacy2} for structural analysis. The results in Table~\ref{tab:eicl} show that fixed template-based generation evidently fails to align with the training distribution. In contrast, our approach preserves structural similarity with the ground-truth distribution, naturally leading to better navigation results when used for on-policy training.

\subsubsection{Semantic Faithfulness of the Hindsight Supervision. }
The proposed TIAW quantifies the semantic fidelity of generated instructions, allowing the framework to dynamically weight the contribution of hindsight supervision based on its reliability. To systematically validate its faithfulness, we measured TIAW across three categories: ground-truth expert references for calibration, zero-shot based, and our Expert-In-Context Learning-based generations derived from on-policy rollouts. 
Table~\ref{tab:tiaw-rel} reports the mean TIAW scores with standard deviation. First, TIAW correctly identifies expert references as having the highest semantic fidelity, establishing a clear upper bound for calibration. Furthermore, Expert-in-context consistently outperforms zero-shot generation in both standard and landmark-aware (TIAW$^{\dagger}$) variants. This provides direct evidence that distributional similarity highly correlates with semantic correspondence between the visual trajectory and the generated text. Altogether, this calibration validates TIAW as a principled filtering mechanism, ensuring that the hindsight supervision is semantically faithful.


\vspace{-1.em}

\begin{table*}[t]
	\centering
	\caption{Distributional alignment between ground-truth and generated instructions.}
    \vspace{-1.0em}
    \label{tab:eicl}
    \resizebox{0.98\textwidth}{!}{
	\begin{tabular}{*{2}{r}|*{3}{c}|*{3}{c}||*{1}{c}}
		\toprule
            &
            & \multicolumn{3}{c|}{Lexical} 
            & \multicolumn{3}{c||}{Structural}
            & \multicolumn{1}{c}{Nav.}\\

		\midrule


        & \textbf{Methods}
        & BLEU-1$\uparrow$ & BLEU-4$\uparrow$  & ROUGE-L$\uparrow$
		& BLEU-1$\uparrow$ & BLEU-4$\uparrow$  & ROUGE-L$\uparrow$&SPL$\uparrow$ \\
        \midrule

        &Zero-shot
        &0.2510&0.022&0.2039    &0.3724&0.1402&0.3492      &50.55 \\
        
        &Template-based~\cite{yan2024instrugen}
        &0.1733&0.0193&0.1648      &0.3201&0.0932&0.2723&- \\
        
        \midrule
        
        &\cellcolor{gray!20}\textbf{Expert-In-Context}
        &\cellcolor{gray!20}\textbf{0.3217}&\cellcolor{gray!20}\textbf{0.087}&\cellcolor{gray!20}\textbf{0.3192}     &\cellcolor{gray!20}\textbf{0.5394}&\cellcolor{gray!20}\textbf{0.3129}&\cellcolor{gray!20}\textbf{0.5952}       &\cellcolor{gray!20}\textbf{52.74}  \\
		\bottomrule
	\end{tabular}
}
\end{table*}

\begin{table}[t]
    \centering
    \caption{Mean TIAW scores with standard deviation}
    \vspace{-1em}
    \begin{tabular}{l|c|cc}
    \toprule
     & Expert Reference & Zero-Shot & \textbf{Expert-In-Context} \\
     \midrule
    TIAW & 0.492 ($\pm$ 0.06) & 0.278 ($\pm$ 0.07) & \textbf{0.307} ($\pm$ 0.07) \\
    TIAW$^{\dagger}$ & 0.517 ($\pm$ 0.07) & 0.291 ($\pm$ 0.05) & \textbf{0.318} ($\pm$ 0.06) \\
    \bottomrule
    \end{tabular}
    \label{tab:tiaw-rel}
    \vspace{-1em}
\end{table}

\subsection{Sample Efficiency Analysis}
\label{sec:sample}
\begin{figure*}[t!]
\centering
\includegraphics[width=\linewidth]{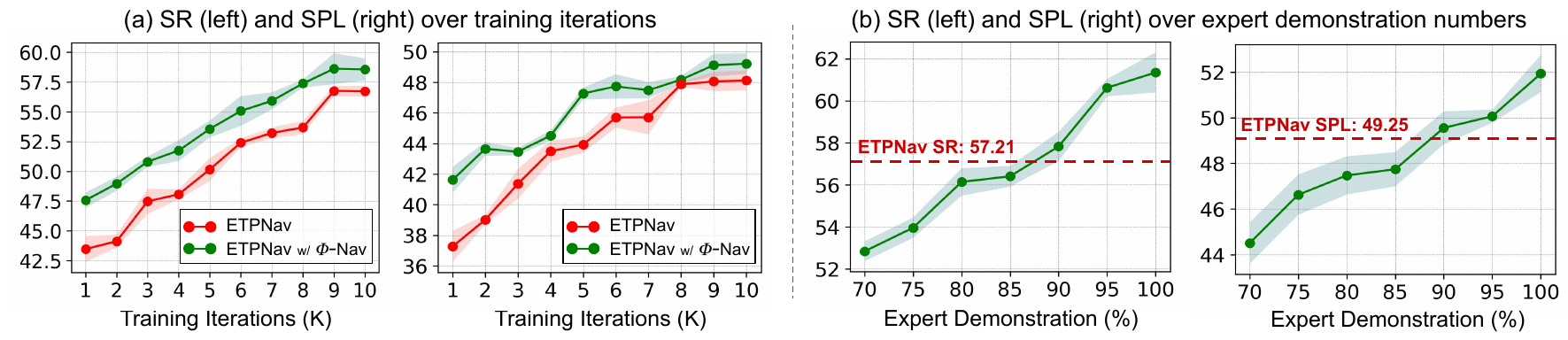}
\vspace{-1.5em}
\caption{\textbf{Sample Efficiency Analysis.} Results are averaged over three independent runs, with the shaded region indicating standard deviation.}
\label{fig:sample}
\end{figure*}

Unlike traditional methods that rely solely on the original expert instructions, $\Phi$-Nav extracts dense learning signals from exploratory deviations, enabling sample efficient training in two perspectives.
First, as illustrated in Figure~\ref{fig:sample}-(a), $\Phi$-Nav consistently outperforms the baseline across all stages of training. This is primarily due to $\Phi$-Nav exposing the agent to a higher volume of state-instruction pairs per iteration. Consequently, $\Phi$-Nav achieves higher success rates with fewer environmental interactions, proving that hindsight supervision is a more potent signal for policy optimization than expert-action correction alone.
Second, Figure~\ref{fig:sample}-(b) highlights $\Phi$-Nav's ability to maintain high performance even when expert demonstrations are scarce. Specifically, we observe that $\Phi$-Nav surpasses the baseline's best result using only 90\% of the available expert data. By augmenting the training distribution with high-fidelity hindsight demonstrations, $\Phi$-Nav enables the baseline model to achieve better performance with a reduced reliance on human-annotated expert data.

\begin{table}[t]
\centering
\setlength{\tabcolsep}{2.3pt} 
\renewcommand{\arraystretch}{0.8} 
\footnotesize

\begin{minipage}{0.44\linewidth}
    \centering
    \caption{Navigation results using different weighting mechanisms. TIAW$^\dagger$ uses landmark extraction.}
    \vspace{-0.6em}
    \label{tab:lambda}
    \begin{tabular}{l|cccc}
        \toprule
        Method & TL & NE$\downarrow$ & SR$\uparrow$ & SPL$\uparrow$ \\
        \midrule
        $\lambda=0.1$ & 13.12 & 4.94 & 59.12 & 48.66 \\
        $\lambda=1.0$ & 13.57 & 5.06 & 58.58 & 46.32 \\
        \midrule
        TIAW  & 12.91 & 4.46  & 61.18 & 51.32 \\
        TIAW$^{\dagger}$ & 12.76 & \textbf{4.29} & \textbf{62.58} & \textbf{52.74} \\
        \bottomrule
    \end{tabular}
\end{minipage}
\hfill
\begin{minipage}{0.50\linewidth}
    \centering
    \caption{Navigation results using LVLM backbones in different sizes.}
    \vspace{-0.6em}
    \label{tab:backbone}
    \begin{tabular}{l|cccc}
        \toprule
        Method & TL & NE$\downarrow$  & SR$\uparrow$ & SPL$\uparrow$ \\
        \midrule
        Qwen2.5-VL-3B & 13.52 & 4.97 & 58.11 & 47.32 \\
        Qwen2.5-VL-7B & 12.76 & \textbf{4.29} & \textbf{62.58} & \textbf{52.74}\\
        Qwen3-VL-8B   & 12.44 & 4.41 & 60.05 & 51.18 \\
        \bottomrule
    \end{tabular}
\end{minipage}
\vspace{-1.5em}
\end{table}

\vspace{-0.5em}
\subsection{Ablation Studies}
\label{sec:ablation}
We provide ablation studies comparing different hindsight weighting mechanisms and LVLM backbones. For these experiments, we use ETPNav~\cite{an2024etpnav} as the baseline and report results on the val unseen split of the R2R-CE dataset.
\vspace{-0.5em}
\subsubsection{Hindsight Weighting Mechanisms. }
In Table~\ref{tab:lambda}, setting the hindsight weight $\lambda$ to a fixed scalar yields suboptimal performance compared to our Trajectory–Instruction Alignment Weighting (TIAW) variants, which adaptively controls the contribution of each sample. Among the fixed-weight settings, smaller $\lambda$ values perform better, suggesting that reducing hindsight signals mitigates learning from noisy or hallucinated supervision.
For the TIAW variant, using the landmark-based scoring consistently outperforms the variant without it across NE, SR, and SPL, indicating that landmark-aware alignment yields more reliable and semantically grounded weighting.
\vspace{-0.5em}
\subsubsection{LVLM Backbones. }
Table~\ref{tab:backbone} reports navigation performance using different LVLM backbones from the Qwen-VL series~\cite{bai2025qwen3}, comparing models of varying sizes. We observe a clear performance gap between the 3B model and the larger variants. Qwen2.5-VL-3B yields the weakest results across all metrics, suggesting that limited model capacity constrains the quality of generated hindsight instructions and downstream policy learning. Comparing Qwen2.5-VL-7B and Qwen3-VL-8B, both larger models achieve substantial improvements over the 3B variant. While Qwen3-VL-8B attains slightly shorter trajectory lengths, Qwen2.5-VL-7B achieves the best overall navigation performance. These results indicate that increasing model scale generally benefits hindsight instruction generation, but architectural differences and alignment characteristics also play a critical role beyond parameter count alone.

\vspace{-0.5em}
\section{Conclusion}
\label{sec:conclusion}
\vspace{-0.5em}
We introduced $\Phi$-Nav, a novel on-policy VLN training framework that addresses the semantic supervision gap arising in exploratory rollouts. Specifically, $\Phi$-Nav converts misaligned trajectories into semantically grounded path-level hindsight instructions, by leveraging the spatio-temporal reasoning of pre-trained LVLMs. Furthermore, to ensure that hindsight supervision remains both faithful and distribution-consistent, we devise expert-in-context learning for structural alignment and trajectory–instruction alignment weighting for adaptive optimization. Experiments on R2R-CE and RxR-CE demonstrate that $\Phi$-Nav not only strengthens standard on-policy imitation learning but also substantially improves sample efficiency, reducing reliance on costly expert demonstrations. Overall, $\Phi$-Nav demonstrates the potential of language-guided self-supervision to advance embodied AI, enabling agents to learn to make decisions from their own experiences.\\

\noindent \textbf{Limitations and Future Work. }
\footnotesize Although our strategies improve instruction generation from LVLMs, the outputs remain imperfect. The trajectory–instruction alignment weight $\lambda$ could also benefit from more sophisticated online, reference-free scoring. Future work will focus on improving generation efficiency, alignment estimation, and extending the framework to longer-horizon and broader robotic VLA tasks.

\section*{Acknowledgement}
This work was supported by Culture, Sports and Tourism R\&D Program through the Korea Creative Content Agency grant funded by the Ministry of Culture, Sports and Tourism
(International Collaborative Research and Global Talent Development for the Development of Copyright Management and Protection Technologies for Generative AI, RS-2024-00345025),
the National Research Foundation of Korea(NRF) grant funded by the Korea government(MSIT)(RS-2025-00521602), Institute of Information \& communications Technology Planning \& Evaluation (IITP) \& ITRC(Information Technology Research Center) grant funded by the Korea government(MSIT) (No.RS-2019-II190079, Artificial Intelligence Graduate School Program(Korea University), 1\%);

\bibliographystyle{splncs04}
\bibliography{main}
\clearpage
\section*{Appendix}
\appendix

\section{Configuration of the Hindsight Speaker Agent}
\subsection{Video Parsing Strategy}
We employ $\Phi$-Nav on two representative VLN baselines: CMA~\cite{krantz2020beyond}, which follows a DAgger-based training scheme, and ETPNav~\cite{an2024etpnav}, which adopts scheduled sampling. In addition to differences in their on-policy exploration strategies, the two baselines use different camera settings: CMA operates in a monocular setting, whereas ETPNav uses panoramic observations. We therefore tailor the visual input for our Hindsight Speaker Agent~(HSA) to match the unique observation requirements of each baseline.

In the monocular setting, each timestep produces a visual observation consisting of a single 224 $\times$ 224 RGB image. The action space is also low-level, including \texttt{FORWARD}, \texttt{TURN LEFT}, \texttt{TURN RIGHT}, and \texttt{STOP}, which results in relatively long visual trajectory sequences. To improve efficiency, we subsample the trajectory by selecting one frame every two timesteps. Additionally, we resize the selected frames to 300 $\times$ 300 so that HSA can capture finer visual details.

In the panoramic setting used in ETPNav, each timestep produces a visual observation consisting of 12 RGB images of size 224 $\times$ 224 with a 90$^\circ$ field of view (FOV), where adjacent views overlap by 60$^\circ$. To reduce redundancy, we select four non-overlapping views centered around the forward-facing direction and concatenate them horizontally to form a 224 $\times$ 896 image representing a 360$^\circ$ panoramic observation for each timestep. Since ETPNav abstracts continuous low-level actions into a topological waypoint-based action space, we use frames from all timesteps to construct the video sequence.

\subsection{Hyperparameters}
Next, we describe the hyperparameters used for HSA, which is based on the Qwen-VL~\cite{bai2025qwen3} large vision-language model (LVLM). We set the temperature to 0.75 to encourage more diverse instruction generation while maintaining coherent outputs. The nucleus sampling parameter is set to top-$p$ = 0.9 and top-$k$ = 50 to allow controlled stochasticity while avoiding low-probability tokens. We enable sampling (\texttt{do\_sample}=True) to encourage diverse yet coherent instruction outputs, and apply a repetition penalty of 1.1 to mitigate redundant phrasing. Finally, key-value caching (\texttt{use\_cache}=True) is enabled to improve efficiency. All other hyperparameters follow the default settings of the respective baseline implementations.

\subsection{Prompt Design}
We present the prompt design used by HSA in Figure~\ref{fig:prompt}. Specifically, we adopt an in-context learning paradigm by providing a trajectory–instruction pair sampled from offline expert demonstrations. This example serves as a reference that guides the model to generate instructions consistent with the distribution of the training data. Consequently, the generated instructions are less likely to introduce noise or abrupt deviations during learning. In addition, we include explicit constraints in the prompt—covering grounding, style, flow, termination, and forbidden expressions—to encourage the pretrained LVLM to produce high-quality and well-structured navigation instructions.

\begin{figure}[t]
\centering
\begin{tcolorbox}[
    colback=gray!5, 
    colframe=NavyBlue, 
    title=\textbf{Expert-In-Context Hindsight Instruction Generation Prompt},
    fonttitle=\small\bfseries,
    fontupper=\small,
    arc=0.5mm
]

\textbf{System:} You are an intelligent assistant that describes videos accurately. \\[0.1em]
\textbf{Expert-In-Context Example:} 
\begin{itemize}[leftmargin=1.2em, noitemsep, topsep=2pt]
    \item \textbf{Video: }\textcolor{WildStrawberry}{\scriptsize \{EXPERT\_TRAJECTORY\_VIDEO\}}
    \item \textbf{User Task: }Based \textit{strictly} on the video sequence of the navigation path, generate a navigation instruction that best describes the path.
    \item \textbf{Constraints:}
        \begin{itemize}[leftmargin=1.0em, noitemsep, topsep=2pt]
        \item \textbf{Grounding:} Refer to visible objects and landmarks in the video.
        \item \textbf{Style:} Use natural and imperative language.
        \item \textbf{Flow:} Combine steps into one fluid paragraph; do not list actions.
        \item \textbf{Termination:} The last sentence must include the word that commands termination of navigation (e.g., `stop', `wait').
        \item \textbf{Forbidden:} Do not use meta-words (e.g., `image', `frame', `camera').
        \end{itemize}
    \item \textbf{Generated Instruction: }\textcolor{WildStrawberry}{\scriptsize \{EXPERT\_TRAJECTORY\_INSTRUCTION\}}
\end{itemize}

\textbf{Main Task:} 
\begin{itemize}[leftmargin=1.2em, noitemsep, topsep=2pt]
    \item \textbf{Video: }\textcolor{Green}{\scriptsize \{ON\_POLICY\_TRAJECTORY\_VIDEO\}}
    \item \textbf{User Task: } <same user task as above>
    \item \textbf{Constraints: } <same constraints as above>
    \item \textbf{Generated Instruction: }
\end{itemize}

\end{tcolorbox}
\vspace{-0.5em}
\caption{Prompt used by the Hindsight Speaker Agent with an expert-in-context example to generate instructions from on-policy trajectories. Note that the <> marks are placeholders for actual texts.}
\label{fig:prompt}
\end{figure}

\section{Distribution of TIAW Across Trajectories}

\begin{figure*}[t!]
\centering
\includegraphics[width=\linewidth]{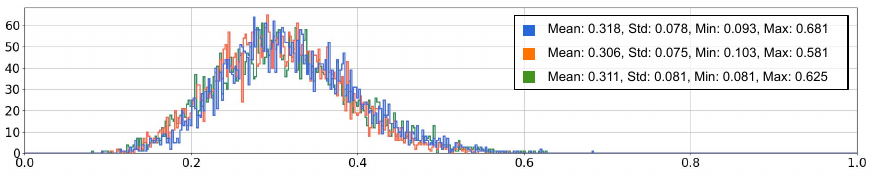}
\vspace{-1.5em}
\caption{Distribution of Trajectory–Instruction Alignment Weights.}
\vspace{-1.5em}
\label{fig:dist}
\end{figure*}

In this section, we analyze the distribution of the Trajectory–Instruction Alignment Weight (TIAW) across trajectories. Specifically, we sample 5,000 distinct on-policy trajectories and compute TIAW across three different random seeds, as shown in Figure~\ref{fig:dist}. Here, the x-axis represents the TIAW values discretized into 500 bins, while the y-axis indicates the corresponding frequency. We report the mean, standard deviation, minimum, and maximum values for each distribution. The results show that the weights are largely concentrated around 0.3, with most values ranging approximately from 0.1 to 0.6. This observation underscores $\Phi$-Nav's ability to adaptively modulate the contribution of hindsight learning signals, assigning greater weight to semantically well-grounded trajectory–instruction pairs while down-weighting less aligned ones, thereby providing more informative and precise supervision for policy training.

\section{Computational Analysis}
\begin{wraptable}{r}{0.43\textwidth} 
\vspace{-2em}   
  \centering
  \begin{tabular}{*{2}{l}|c|c}
    \hline
    &Method & TPS$\uparrow$ & APS$\uparrow$\\ \hline
    &CMA-D~\cite{krantz2020beyond}    & 0.66    & 32.22\\
    &\multicolumn{1}{r|}{w/ $\Phi$-Nav}     & \textbf{0.71}    & \textbf{34.64} \\ \hline
  \end{tabular}
  \vspace{-0.5em}
  \caption{Computational Analysis}
  \label{tab:comp}
  \vspace{-2em}
\end{wraptable}
We evaluate the computational efficiency of $\Phi$-Nav in the DAgger-based setting by measuring the time required to collect trajectory–instruction pairs for supervision. Specifically, we report two metrics: \textit{trajectory-instruction per second} (TPS), which quantifies the average number of on-policy trajectories paired with language instructions generated per second, and \textit{action-instruction per second} (APS), which measures the average number of action–instruction pairs generated for those trajectories per second. APS can be interpreted as a step-wise version of TPS, providing a finer-grained view of computational throughput. As reported in Table~\ref{tab:comp}, $\Phi$-Nav enables the policy to collect a larger amount of supervision pairs within a given time budget of total 600 seconds. We acknowledge the inevitable increase in wall-clock training time due to the additional forward pass of the LVLM for instruction generation. However, this overhead is relatively minor when considering the substantial increase in the number of semantically grounded supervision obtained for policy learning.

%

\section{Additional Discussions}
\vspace{-0.5em}
In this section, we highlight several notable questions and insights that emerged during the research process.\\
\vspace{-0.5em}

\noindent \textbf{Question 1: How is $\Phi$-Nav different from Speaker-Follower models?} \\
\noindent \textbf{Answer: } While $\Phi$-Nav naturally augments the set of training episodes, the contribution differs fundamentally from prior Speaker-Follower (SF)-based augmentation methods. First, the primary objective of SF is to increase the diversity of offline trajectory-instruction pairs, whereas $\Phi$-Nav aims to address the semantic mismatch that arises during on-policy exploration. Furthermore, while SF generates instructions for static offline trajectories, $\Phi$-Nav instead generates hindsight instructions conditioned on the agent’s own exploratory trajectories. Consequently, $\Phi$-Nav introduces challenges beyond conventional offline augmentation, including dynamically relabeling sub-optimal on-policy trajectories and preserving semantic consistency in hindsight supervision. Our proposed Expert-in-context Learning and TIAW, respectfully, are specifically designed to address these challenges, which do not arise in SF.\\

\noindent \textbf{Question 2: Can $\Phi$-Nav be applied in other embodied navigation tasks?} \\
\noindent \textbf{Answer: } We believe that other embodied navigation tasks, such as object-goal navigation~\cite{chaplot2020object} or vision-based navigation~\cite{shah2021ving}, could potentially benefit from the idea of learning from hindsight experiences. However, VLN presents a particularly challenging setting due to the need to ground dense, long-horizon language instructions within complex visual environments. $\Phi$-Nav is primarily designed for such scenarios by leveraging hindsight instruction generation to bridge the semantic supervision gap in on-policy training. Therefore, while the underlying principle of hindsight-based learning may extend to general embodied navigation tasks, $\Phi$-Nav specifically addresses the unique challenges of semantic grounding and instruction following in language-driven navigation, which more closely resembles real-world human–robot interaction. 
\\

\noindent \textbf{Question 3: Does $\Phi$-Nav improve error recovery behavior?} \\
\noindent \textbf{Answer: }Error recovery is a critical capability for embodied navigation agents, particularly in unseen environments where deviations from the optimal path frequently occur. While $\Phi$-Nav does not explicitly introduce a dedicated mechanism for error recovery, it indirectly supports this behavior through improved semantic supervision during training. Specifically, $\Phi$-Nav increases the density of semantically grounded trajectory–instruction pairs by generating hindsight instructions for on-policy exploratory trajectories. This enriched supervision expands the agent’s semantic state coverage, enabling the policy to better associate environmental states with language-guided navigation behaviors. As a result, the agent becomes more capable of recognizing and correcting deviations during navigation, which contributes to improved performance in unseen environments and yields a positive effect on error recovery.\\

\noindent \textbf{Question 4: What new research directions does $\Phi$-Nav suggest?} \\
\noindent \textbf{Answer: }$\Phi$-Nav opens a pathway for embodied agents to self-analyze and reason over their own experiences. Considering both the contributions and limitations, we identify the following topics as promising directions for future research:
\footnotesize
\begin{enumerate}
    \item \textbf{Reducing Reliance on Offline Expert Demonstrations.} \\
    As demonstrated in Section 4.4 of the main manuscript, $\Phi$-Nav achieves competitive performance compared to the baseline while using approximately 10\% less offline expert demonstration data. This finding suggests a promising step toward mitigating the data-hungry nature of VLN training. Further efforts to reduce reliance on expert demonstrations could benefit not only VLN but also instruction-following embodied policy learning more broadly.\\
    
    \item \textbf{Improving Trajectory-Instruction Alignment Scoring.}  \\
    Just as generating high-quality instructions from trajectory videos is important, developing reliable numerical measures to assess trajectory–instruction alignment is equally critical. In this work, we employ a lightweight scoring mechanism inspired by EMScore~\cite{shi2022emscore}. However, developing more robust alignment metrics could further improve the effectiveness of hindsight supervision. Future work may explore leveraging 3D vision–language foundation models for better spatial grounding, as well as sequence-level alignment between visual trajectories and linguistic structures.\\

    \item \textbf{Enabling Intermediate Hindsight Reasoning.}\\
    $\Phi$-Nav enables embodied agents to retrospectively analyze their exploratory trajectories once navigation is completed. While this effectively bridges the semantic supervision gap in on-policy training, the training process could further benefit from intermediate hindsight reasoning during navigation, rather than only after termination. Such intermediate reflection may enable finer-grained path understanding across trajectories of varying lengths. However, introducing intermediate reasoning also raises challenges, including increased latency and the need to maintain spatial and temporal consistency throughout the trajectory, making this a challenging yet promising direction for future research.
    
\end{enumerate}


\end{document}